\documentclass{article} 
\usepackage{colm2024_conference}
\usepackage{graphicx}
\usepackage{microtype}
\usepackage{hyperref}
\usepackage{url}
\usepackage{booktabs}
\definecolor{darkblue}{rgb}{0, 0, 0.5}
\hypersetup{colorlinks=true, citecolor=darkblue, linkcolor=darkblue, urlcolor=darkblue}
\usepackage{xspace}
\usepackage{amssymb}
\usepackage{amsmath}
\usepackage{algorithm}
\usepackage{algorithmic}
\usepackage{wrapfig}
\usepackage{nicematrix}
\usepackage{tablefootnote}
\usepackage{amssymb}
\usepackage{pifont}
\usepackage{adjustbox}
\usepackage{todonotes}
\usepackage{csquotes}
\usepackage{tikzpagenodes}

\newcommand{\lm}{Granite Code models\xspace}

\usepackage{caption}
\usepackage{subcaption}
\usepackage{algorithm}
\usepackage{algorithmic}
\usepackage{multirow}
\usepackage{xcolor}
\usepackage{fontawesome}

\makeatletter
\newcommand{\makecmd}[2]{%
  \expandafter\newcommand\csname #1\endcsname[1]{%
    #2%
    \def\tempa{##1}%
    \ifx\tempa\@empty
    \else
      -##1B%
    \fi
  }%
}
\makeatother

\makecmd{starcodertwo}{StarCoder2}
\makecmd{starcoderbase}{StarCoderBase}
\makecmd{deepseekcoder}{DeepSeekCoder}
\makecmd{stablecode}{StableCode}
\makecmd{codellama}{CodeLlama}
\makecmd{octocoder}{OctoCoder}
\makecmd{gemma}{Gemma}
\makecmd{codegemma}{CodeGemma}
\makecmd{llamathree}{Llama-3}
\makecmd{mistral}{Mistral}
\makecmd{wizardcoder}{WizardCoder}
\makecmd{lemma}{Lemma}

\makecmd{codellamainstruct}{CodeLlama-Instruct}
\makecmd{deepseekcoderinstruct}{DeepSeekCoder-Instruct}

\title{\centering \LARGE Scaling Granite Code Models to 128K Context}

\colmfinalcopy 
\begin{document}
\maketitle

\vspace{0.5cm}
\vspace{-2cm}
\begin{center}

\textbf{Matt~Stallone}\quad
\textbf{Vaibhav Saxena}\quad
\textbf{Leonid Karlinsky}\quad
\textbf{Bridget McGinn}\quad
\textbf{Tim Bula}\quad
\textbf{Mayank~Mishra}\quad
\textbf{Adriana~Meza~Soria}\quad
\textbf{Gaoyuan~Zhang}\quad
\textbf{Aditya~Prasad}\quad
\textbf{Yikang~Shen}\quad
\textbf{Saptha~Surendran}\quad
\textbf{Shanmukha Guttula}\quad
\textbf{Hima~Patel}\quad
\textbf{Parameswaran Selvam}\quad
\textbf{Xuan-Hong~Dang}\quad
\textbf{Yan~Koyfman}\quad
\textbf{Atin Sood}\quad
\textbf{Rogerio Feris}\quad\\
\textbf{Nirmit~Desai}\quad
\textbf{David~D.~Cox}\quad
\textbf{Ruchir~Puri}$^\dagger$\quad
\textbf{Rameswar~Panda}$^\dagger$\quad

IBM Research\\
 $^\dagger$Corresponding Authors \\
 \texttt{ruchir@us.ibm.com, rpanda@ibm.com}

\end{center}

\begin{abstract}
This paper introduces long-context Granite code models that support effective context windows of up to 128K tokens. Our solution for scaling context length of Granite 3B/8B code models from 2K/4K to 128K consists of a light-weight continual pretraining by gradually increasing its RoPE base frequency with repository-level file packing and length-upsampled long-context data. Additionally, we also release instruction-tuned models with long-context support which are derived by further finetuning the long context base models on a mix of permissively licensed short and long-context instruction-response pairs. While comparing to the original short-context Granite code models, our long-context models achieve significant improvements on long-context tasks without any noticeable performance degradation on regular code completion benchmarks (e.g., HumanEval). We release all our long-context \lm under an Apache 2.0 license for both research and commercial use.  

\ifcolmfinal
\vspace{.75em}
\centering \faGithubSquare~ \url{https://github.com/ibm-granite/granite-code-models}
\else
\fi

\end{abstract}

\section{Introduction}
\label{sec:introduction}
With the emergence and development of repository-level coding tasks~\citep{repoqa,liu2023repobench} and software development agents~\citep{opendevin2024}, long context length becomes an important feature for code language models. 
While many proprietary large language models, like GPT4, Gemini, and Claude, support very long context windows, most open-source code language models could only provide relatively short context windows~\citep{codegemma,codellama}.
This short context length limits the practicality of open-source code language models in real-world software development.

In this paper, we introduce the long-context Granite code 3B and 8B, a series of code language models that support effective context lengths up to 128K tokens.
To achieve the extended context length, we first continue pretrain Granite Code 3B/8B base models with a repository-level code corpus and upsample the longer context repositories.
Then, we instruction tune the continue pretrained model on a combination of short and long context instruction data. 
Due to the lack of long context instruction data, we generate multi-turn instruction data from repository-level file-packed documents with our original Granite-8B-Code-Instruct model to avoid the dependency on an existing long context model. 
More details of long context extension can be found in Section~\ref{sec:long context}.

To evaluate the ability of long-context Granite Code models, we conduct extensive experiments on both short and long-context tasks, including HumanEvalPack, Long Code Completion, RepoBench-P, RepoQA, and Key Retrieval. 
Experiment results show that our long-context models significantly improve long-context performances without noticeable degradation in short-context performances.
We open-source all our long-context Granite Code models under an Apache 2.0 license for research and commercial use.

\section{Long Context Modeling}
\label{sec:long context}

Our solution for scaling context length of Granite code models consists of a continual pretraining and an instruction tuning phase. Similar to prior works~\citep{fu2024data}, we hold the basic hypothesis that the ability to utilize information at arbitrary input locations, is a capability that is mostly already acquired through large-scale pretraining,
and that this capability can be readily extended
to contexts substantially longer than seen during
original pretraining (e.g., 4K to 128K) through lightweight training on appropriate data mixture.

\subsection{Continual Pretraining}
We continue pretrain the full attention Granite code base models using sequence parallelism\footnote{\url{https://github.com/jzhang38/EasyContext}}~\citep{li2021sequence} by gradually increasing its RoPE base frequency without using any sparse or linear attention.
Specifically, we continue pretrain Granite Code 3B/8B base models using the original pretraining data used in~\cite{mishra2024granite} but with repository-level file packing and per-language context length upsampling, that we found to be critical for long-context continual pretraining. This continued training stage focused on a curated selection of programming languages, such as Python, C, C++, Go, Java, JavaScript, and TypeScript, as in~\cite{pinnaparaju2024stable}.

To create long-context data, we develop a new approach that packs files from the same repository together, arranging them to prioritize semantic dependencies. We identify these dependencies by analyzing file imports and create a directed acyclic graph, where each file is a node and edges represent API imports between files. After breaking any cycles in the graph, we perform a topological sort to establish an ordering of files based on their semantic dependencies. We then organize the files in a repository by placing documentation and build files first, followed by the ordered set of files with semantic dependencies, and finally the remaining non-connected files. These non-connected files are arranged according to their folder structure, using a depth-first search to traverse the repository. Finally, we determine the dominant programming language of a repository based on file extensions and presence of build files, to organise repo-ordered files by programming languages.

The documents’ lengths and their source domains/languages are two closely related confounding factors in data engineering because long data usually come from particular sources.
Thus, in addition to repository-level file packing, we artificially oversampled longer document sequences on a per-language basis to ensure the quantity of long sequences, thereby improving the overall quality of our training data corpus, as in ~\cite{fu2024data,yu2023paraphrasing}. In particular, we downsample documents under 4096 tokens to a rate of 10\%, which we find to ensure a sufficient number of total tokens and documents. The total number of documents within the training corpus after processing is 173,336 with a mean length of 73,451.

We adjust the RoPE base frequency, introduced
in \cite{xiong2023effective}, to support long context windows up to 128K where the base model itself is trained on 2K/4K context length.
For training, we adopt a progressive approach where we doubled the context window until it reached the desired length of 128K.
We train for 500 steps with a batch size of 32 and search for the optimal RoPE theta and learning rate for each iteration. For RoPE theta, we finf optimal values of 100K, 250K, 500K, 2M, and 10M for context windows of 8K, 16K, 32K, 64K, and 128K, respectively. We train with data parallelism and Flash Attention 2 until 64K tokens and then used Ring Attention ~\citep{liu2023ringattentionblockwisetransformers} to reach 128K tokens. The final models are trained for an extra 4B tokens which is only 0.1\% of original pretraining data. 

\subsection{Instruction Tuning}

Our training data for long context instruct models consists of a combination of permissively licensed data used in training the original Granite code instruct models~\citep{mishra2024granite}, in addition to synthetically generated code instruction datasets tailored for solving long context problems. Specifically, the 128K long context instruct models are derived by
further finetuning the long context base models on a mix of short and long context data as follows.

\textbf{Short-Context Instruction Data.} Our short context instruction data consists of  a combination of CommitPackFT~\citep{octopack}, MathInstruct\footnote{We removed GSM8K-RFT and Camel-Math from MathInstruct due to unknown or NC license.}~\citep{yue2023mammoth}, MetaMathQA~\citep{yu2023metamath}, Glaive-Code-Assistant-v3\footnote{\url{https://huggingface.co/datasets/glaiveai/glaive-code-assistant-v3}}, Self-OSS-Instruct-SC2\footnote{\url{https://huggingface.co/datasets/bigcode/self-oss-instruct-sc2-exec-filter-50k}}, Glaive-Function-Calling-v2\footnote{\url{https://huggingface.co/datasets/glaiveai/glaive-function-calling-v2}}, NL2SQL\footnote{\url{https://huggingface.co/datasets/bugdaryan/sql-create-context-instruction}}, HelpSteer~\citep{wang2023helpsteer}, OpenPlatypus\footnote{\url{https://huggingface.co/datasets/garage-bAInd/Open-Platypus}}~\citep{platypus2023}, and a few synthetically generated datasets for API calling ~\citep{basu2024api}, and multi-turn code interactions with execution feedback. 

\textbf{Long-Context Instruction Data.} The long context instruction data was synthetically generated by bootstrapping the pretraining data. For each repository-level file-packed document, we created a multi-turn dataset where the instructions within each sample were human-designed for the purpose of enhancing the long-context performance in specific tasks like generation, retrieval and translation. The responses were either parsed semantically from the original document or generated using Granite-8b-Code-Instruct-4K. 
The dataset first parses the document into classes, methods, and stand-alone functions. It then requests and extracts the implementations of a random subset of the extracted functions/methods (up to 5 per file in the document) and then asks for an explanation of that implementation using available documentation. Additionally, it generates instructions for implementing the sampled functions (methods) based on the remaining documentation and code with the function excluded.
These questions and instructions were repeated for different functions until the desired length was achieved.

By exposing the model to both short and long context data, we aim to enhance its long context capability without sacrificing code generation performance at short input context. For finetuning, we use a multiturn loss mask for each sample, as in \cite{wang2023farcamelsgoexploring}. This is particularly important as our finetuning data corpus consists of instruction-response pairs with multiple turns. However, when composing a sequence, we append an \texttt{EOS} token after each response from the model to prevent runaway generation during inference. We followed the same training parameters that produced our previous short-context instruct models~\citep{mishra2024granite}: 128 global batch size, 2e-5 learning rate, a noise multiplier of 5 for input embeddings, and padding-free transformers. 

\section{Results}
\label{sec:results}

We evaluate our long-context Granite code models on a wide variety of benchmarks by measuring key retrieval accuracy and performance during generation on code completion tasks at both short and long-context length as follows.

\subsection{Benchmarks}

\textbf{Long Code Completion.} Long Code Completion (LCC) \citep{guo2023longcoderlongrangepretrainedlanguage} tests a model's ability to predict the next line of code from long repository-based context for Python, Java, and C\#. While the benchmark's context length spans 1/2K through 8K+ tokens, it is heavily weighted around 2K tokens. Thus, following \cite{bai2024longbenchbilingualmultitaskbenchmark} and \cite{codellama}, we rebalance this dataset for equal representation with each context length bucket (\textless4K, 2 \textendash\;4K, 4 \textendash\;8K, 8K+), where each bucket has 100 samples when possible. 

\begin{table}[h]
    \caption{Exact Match (EM) performance on Long Code Completion (LCC) benchmark (Balanced). Long-context Granite code models consistently outperforms original base models at different input context from 4K to 32K.}
    \label{tab:lcc}
    \centering
    \small
    \begin{tabular}{c|cccc}
        \toprule
        \textbf{Model} & \textbf{4K EM} & \textbf{8K EM} & \textbf{16K EM} & \textbf{32K EM} \\
        \midrule
        Granite-3b-Code-Base-2K & 24.5 & 15.4 & 11.4 & 10.0 \\
        Granite-3b-Code-Base-128K & 54.6 & 56.8 & 52.2 & 57.8 \\
        \midrule
        Absolute Gap & + 30.1 & + 41.4 & + 40.8 & + 47.8 \\
        \toprule
        Granite-8b-Code-Base-4K & 41.9 & 23.7 & 19.1 & 15.0 \\
        Granite-8b-Code-Base-128K & 56.5 & 60.1 & 51.8 & 57.4 \\
        \midrule        
        Absolute Gap & + 14.6 & + 36.4 & + 32.7 & + 42.4 \\
       \midrule 
    \end{tabular}
\end{table}

\textbf{RepoBench-P.} Like LCC, RepoBench-P \citep{liu2023repobenchbenchmarkingrepositorylevelcode} tests the model’s next line code completion ability for long-context input. We follow the methodology in \citep{bai2024longbenchbilingualmultitaskbenchmark} by selecting the Cross-File-First data but then we rebalance the buckets based on the Starcoder tokenizer used for training out Granite code models.

\textbf{RepoQA.} RepoQA~\citep{repoqa} is an advanced Needle-in-the-Haystack test that focuses on testing LLMs' capabilities on long-context code understanding and retrieval.
Specifically, given a long chunk of source code and a precise function description, and the model is asked to find the function in the context that corresponds to the description. This benchmark focuses on retrieving 10 needle functions from each of 5 languages x 10 repositories (500 sub-tasks/tests) with a set context size of 16K tokens. 

\begin{table}
    \caption{Exact Match (EM) scores on RepoBench-P (Balanced) benchmark.}
    \label{tab:repobench-p}
    \centering
    \small
    \begin{tabular}{c|cccc}
        \toprule
        \textbf{Model} & \textbf{4K EM} & \textbf{8K EM} & \textbf{16K EM} & \textbf{32K EM} \\
        \midrule
        Granite-3b-Code-Base-2K & 22.0 & 17.9 & 15.4 & 14.0 \\
        Granite-3b-Code-Base-128K & 39.8 & 46.8 & 43.1 & 45.3 \\
        \midrule
        Absolute Gap & + 17.8 & + 28.9 & + 27.7 & + 31.3 \\
        \toprule
        Granite-8b-Code-Base-4K & 27.9 & 23.0 & 15.7 & 7.8 \\
        Granite-8b-Code-Base-128K & 42.7 & 44.0 & 44.8 & 44.5 \\
        \midrule
        Absolute Gap & + 14.8 & + 21.0 & + 29.1 & + 36.7 \\
        \bottomrule
    \end{tabular}
\end{table}

\textbf{Key Retrieval.} This is a synthetic benchmark that tests the model’s ability to find and execute a Python function buried within high-quality, syntactically correct Python code. As proposed in \cite{codellama}, we took the Code Contest finetuning dataset from \cite{alphacoder} and concatenated Python solutions around the key function. We then asked the model to return the output of the key function by emulating a Python interpreter shell. We created sequences of lengths of 512 tokens and key offsets of 512 tokens. 

\textbf{HumanEvalPack.} To evaluate model performance at short-context length, we adopt HumanEvalPack~\citep{octopack}, which extends Python problems of Humaneval Benchmark to five additional commonly used programming languages, namely JavaScript, Java, Go, C++, Rust to test three coding tasks (generation, explanation and fixing). We evaluate our long-context models in a zero-shot manner using greedy decoding
with completion format for the base models, and with instruction template for the instruction-tuned models.

\subsection{Base Model Evaluations} 
Table~\ref{tab:lcc} and Table~\ref{tab:repobench-p} show the results of Granite 3B/8B code models before and after long-context extension on LCC and RepoBench-P benchmarks respectively. Prior Granite code models with 2K/4K support fail to generate meaningful completions on long sequences. On the other hand, across all the context length (4K to 32K), models scaled to handle long contexts up to 128K achieve significantly higher performance. This demonstrates that
long contexts are informative for code completion, and long-context Granite code models are able to effectively leverage this information to improve their generations on both benchmarks.

In Table~\ref{tab:repoqa-base}, we
compare the performance of Granite code base models to their counterparts prior to long-context
extension. Our long-context models exhibit strong retrieval performance across different matching thresholds, while the short context versions mostly fail in finding the needle function successfully. The absolute differences averaged over 5 programming languages are very significant, e.g., $+38.6\%$ for Granite 8B model with a matching threshold of 0.8.
By looking at the score
distribution across different programming languages, we can
see that both models are doing best at Python, with 8B model consistently outperforming the 3B model. This result shows that our long-context Granite code models can better understand natural language description before retrieval, which aligns with the use of advanced code search in many practical situations.


\begin{table}[t]
    \caption{Retrieval accuracy (\%) of Granite code base models on RepoQA benchmark evaluated using 16K context length at multiple thresholds of match similarity. All models are evaluated using greedy decoding with 256 new token limit.}
    \label{tab:repoqa-base}
    \centering
    \small
    \resizebox{\linewidth}{!}{\begin{tabular}{c|ccccccccccc}
        \toprule
        \textbf{Threshold} & \textbf{0.0} & \textbf{0.1} & \textbf{0.2} & \textbf{0.3} & \textbf{0.4} & \textbf{0.5} & \textbf{0.6} & \textbf{0.7} & \textbf{0.8} & \textbf{0.9} & \textbf{1.0} \\
        \midrule
        \multicolumn{12}{c}{\textbf{Granite-3b-Code-Base-2K}} \\
        \midrule
        Python & 6.0  & 0.0 & 0.0 & 0.0 & 0.0 & 0.0 & 0.0 & 0.0 & 0.0 & 0.0 & 0.0 \\
        C++ & 6.0  & 0.0 & 0.0 & 0.0 & 0.0 & 0.0 & 0.0 & 0.0 & 0.0 & 0.0 & 0.0 \\
        Java & 4.0  & 0.0 & 0.0 & 0.0 & 0.0 & 0.0 & 0.0 & 0.0 & 0.0 & 0.0 & 0.0 \\
        TypeScript & 7.0  & 0.0 & 0.0 & 0.0 & 0.0 & 0.0 & 0.0 & 0.0 & 0.0 & 0.0 & 0.0 \\ 
        Rust & 1.5 & 0.0 & 0.0 & 0.0 & 0.0 & 0.0 & 0.0 & 0.0 & 0.0 & 0.0 & 0.0 \\
        \textbf{Average} & 4.9  & 0.0 & 0.0 & 0.0 & 0.0 & 0.0 & 0.0 & 0.0 & 0.0 & 0.0 & 0.0 \\
        \midrule
        \multicolumn{12}{c}{\textbf{Granite-3b-Code-Base-128K}} \\
        \midrule
        Python & 76.0 & 57.0 & 54.0 & 49.0 & 44.0 & 40.0 & 34.0 & 30.0 & 28.0 & 25.0 & 20.0 \\
        C++ & 58.0 & 48.0 & 44.0 & 41.0 & 39.0 & 36.0 & 33.0 & 31.0 & 30.0 & 24.0 & 17.0 \\
        Java & 59.0 & 50.0 & 44.0 & 42.0 & 40.0 & 37.0 & 35.0 & 31.0 & 26.0 & 20.0 & 16.0 \\
        TypeScript & 58.0 & 38.0 & 34.0 & 33.0 & 29.0 & 27.0 & 23.0 & 23.0 & 23.0 & 16.0 & 7.0 \\
        Rust & 57.0 & 38.0 & 36.0 & 32.0 & 30.0 & 29.0 & 28.0 & 24.0 & 24.0 & 19.0 & 16.0 \\
        \textbf{Average} & 61.6 & 46.2 & 42.4 & 39.4 & 36.4 & 33.8 & 30.6 & 27.8 & 26.2 & 20.8 & 15.2 \\
        \midrule
        \textbf{Absolute Gap} & + 56.7 & + 46.2 & + 42.4 & + 39.4 & + 36.4 & + 33.8 & + 30.6 & + 27.8 & + 26.2 & + 20.8 & + 15.2 \\
        \toprule
        \multicolumn{12}{c}{\textbf{Granite-8b-Code-Base-4K}} \\
        \midrule
        Python& 9.0  & 1.0 & 1.0 & 1.0 & 1.0 & 1.0 & 1.0 & 1.0 & 1.0 & 1.0 & 1.0 \\
        C++ & 10.0 & 2.0 & 2.0 & 2.0 & 2.0 & 2.0 & 2.0 & 2.0 & 2.0 & 1.0 & 1.0 \\
        Java & 11.0 & 1.0 & 1.0 & 1.0 & 1.0 & 0.0 & 0.0 & 0.0 & 0.0 & 0.0 & 0.0 \\
        TypeScript & 9.0  & 0.0 & 0.0 & 0.0 & 0.0 & 0.0 & 0.0 & 0.0 & 0.0 & 0.0 & 0.0 \\
        Rust & 11.0 & 0.0 & 0.0 & 0.0 & 0.0 & 0.0 & 0.0 & 0.0 & 0.0 & 0.0 & 0.0 \\
        \textbf{Average} & 10.0 & 0.8 & 0.8 & 0.8 & 0.8 & 0.6 & 0.6 & 0.6 & 0.6 & 0.4 & 0.4 \\
        \midrule
        \multicolumn{12}{c}{\textbf{Granite-8b-Code-Base-128K}} \\
        \midrule
        Python & 85.0 & 73.0 & 69.0 & 68.0 & 66.0 & 65.0 & 62.0 & 58.0 & 54.0 & 51.0 & 45.0 \\
        C++ & 60.0 & 45.0 & 42.0 & 40.0 & 37.0 & 35.0 & 35.0 & 34.0 & 32.0 & 27.0 & 23.0 \\
        Java & 57.0 & 52.0 & 48.0 & 44.0 & 42.0 & 39.0 & 38.0 & 36.0 & 32.0 & 28.0 & 23.0 \\
        Typescript & 64.0 & 55.0 & 49.0 & 48.0 & 44.0 & 40.0 & 38.0 & 36.0 & 35.0 & 28.0 & 12.0 \\
        Rust & 74.0 & 67.0 & 65.0 & 59.0 & 57.0 & 54.0 & 51.0 & 46.0 & 43.0 & 38.0 & 31.0 \\
        \textbf{Average} & 68.0 & 58.4 & 54.6 & 51.8 & 49.2 & 46.6 & 44.8 & 42.0 & 39.2 & 34.4 & 26.8 \\
        \midrule
        \textbf{Absolute Gap} & + 58.0 & + 57.6 & + 54.8 & + 51.0 & + 48.6 & +46.0 & + 44.2 & + 41.4 & + 38.6 & + 34.0 & + 26.4 \\
        \bottomrule
    \end{tabular}}
\end{table}
\begin{table}[t]
    \caption{Retrieval accuracy (\%) of Granite code instruct models on RepoQA benchmark at different matching thresholds (larger represent closer to exact match).}
    \label{tab:repoqa-ins}
    \centering
    \small
    \resizebox{\linewidth}{!}{\begin{tabular}{c|ccccccccccc}
        \toprule
        \textbf{Threshold} & \textbf{0.0} & \textbf{0.1} & \textbf{0.2} & \textbf{0.3} & \textbf{0.4} & \textbf{0.5} & \textbf{0.6} & \textbf{0.7} & \textbf{0.8} & \textbf{0.9} & \textbf{1.0} \\
        \midrule
        \multicolumn{12}{c}{\textbf{Granite-3b-Instruct-Base-2K}} \\
        \midrule
        Python & 15.0  & 0.0  & 0.0  & 0.0  & 0.0  & 0.0  & 0.0  & 0.0  & 0.0  & 0.0  & 0.0 \\
        C++ & 10.0  & 1.0  & 1.0  & 1.0  & 1.0  & 1.0  & 0.0  & 0.0  & 0.0  & 0.0  & 0.0 \\
        Java & 8.0   & 0.0  & 0.0  & 0.0  & 0.0  & 0.0  & 0.0  & 0.0  & 0.0  & 0.0  & 0.0 \\
        TypeScript & 11.0  & 0.0  & 0.0  & 0.0  & 0.0  & 0.0  & 0.0  & 0.0  & 0.0  & 0.0  & 0.0 \\ 
        Rust & 9.0   & 0.0  & 0.0  & 0.0  & 0.0  & 0.0  & 0.0  & 0.0  & 0.0  & 0.0  & 0.0 \\
        \textbf{Average} & 10.6  & 0.2  & 0.2  & 0.2  & 0.2  & 0.2  & 0.0  & 0.0  & 0.0  & 0.0  & 0.0 \\
        \midrule
        \multicolumn{12}{c}{\textbf{Granite-3b-Code-Instruct-128K}} \\
        \midrule
        Python & 76.0 & 60.0 & 55.0 & 54.0 & 50.0 & 48.0 & 42.0 & 41.0 & 40.0 & 38.0 & 33.0 \\
        C++ & 58.0 & 48.0 & 44.0 & 41.0 & 39.0 & 36.0 & 33.0 & 31.0 & 30.0 & 24.0 & 17.0 \\
        Java & 59.0 & 51.0 & 43.0 & 42.0 & 40.0 & 38.0 & 35.0 & 31.0 & 26.0 & 21.0 & 19.0 \\
        TypeScript & 80.0 & 68.0 & 54.0 & 50.0 & 43.0 & 39.0 & 36.0 & 35.0 & 29.0 & 20.0 & 9.0 \\
        Rust & 67.0 & 44.0 & 36.0 & 33.0 & 32.0 & 29.0 & 28.0 & 26.0 & 24.0 & 20.0 & 16.0 \\
        \textbf{Average} & 68.0 & 54.0 & 46.4 & 44.0 & 42.6 & 38.0 & 34.8 & 32.8 & 29.8 & 24.6 & 18.8 \\
        \midrule
        \textbf{Absolute Gap} & + 77.4 & + 53.8 & + 46.2 & + 43.8 & + 42.4 & + 37.8 & + 34.8 & + 32.8 & + 29.8 & + 24.6 & + 18.8 \\
        \toprule
        \multicolumn{12}{c}{\textbf{Granite-8b-Code-Instruct-4K}} \\
        \midrule
        Python & 3.0  & 2.0 & 1.0 & 0.0 & 0.0 & 0.0 & 0.0 & 0.0 & 0.0 & 0.0 & 0.0 \\
        C++ & 10.0 & 2.0 & 2.0 & 2.0 & 2.0 & 2.0 & 2.0 & 2.0 & 2.0 & 2.0 & 1.0 \\
        Java & 8.0  & 1.0 & 1.0 & 1.0 & 1.0 & 1.0 & 1.0 & 1.0 & 1.0 & 1.0 & 0. \\
        TypeScript &  10.0 & 0.0 & 0.0 & 0.0 & 0.0 & 0.0 & 0.0 & 0.0 & 0.0 & 0.0 & 0.0 \\ 
        Rust & 4.0  & 0.0 & 0.0 & 0.0 & 0.0 & 0.0 & 0.0 & 0.0 & 0.0 & 0.0 & 0.0 \\
        \textbf{Average} & 7.0  & 1.0 & 0.8 & 0.6 & 0.6 & 0.6 & 0.6 & 0.6 & 0.6 & 0.6 & 0.2 \\
        \midrule
        \multicolumn{12}{c}{\textbf{Granite-8b-Code-Instruct-128K}} \\
        \midrule
        Python & 89.0 & 83.0 & 81.0 & 79.0 & 76.0 & 73.0 & 67.0 & 63.0 & 58.0 & 52.0 & 48.0 \\
        C++ & 63.0 & 51.0 & 46.0 & 42.0 & 41.0 & 37.0 & 36.0 & 30.0 & 24.0 & 15.0 & 3.0 \\
        Java & 91.0 & 84.0 & 79.0 & 77.0 & 76.0 & 73.0 & 69.0 & 66.0 & 63.0 & 46.0 & 39.0 \\
        TypeScript & 86.0 & 84.0 & 80.0 & 72.0 & 68.0 & 62.0 & 56.0 & 49.0 & 40.0 & 25.0 & 11.0 \\ 
        Rust & 83.0 & 78.0 & 73.0 & 67.0 & 65.0 & 63.0 & 60.0 & 55.0 & 53.0 & 48.0 & 40.0 \\
        \textbf{Average} & 82.4 & 76.0 & 71.8 & 67.4 & 65.2 & 61.6 & 57.6 & 52.6 & 47.6 & 37.2 & 28.2  \\
        \midrule
        \textbf{Absolute Gap} & + 75.4 & + 75.0 & + 71.0 & + 66.8 & + 64.6 & + 61.0 & + 57.0 & + 52.0 & + 47.0 & + 36.6 & + 28.0 \\
        \bottomrule
    \end{tabular}}
\end{table}

\subsection{Instruct Model Evaluations}

\begin{figure}[t]
\centering
\includegraphics[width=0.8\textwidth]{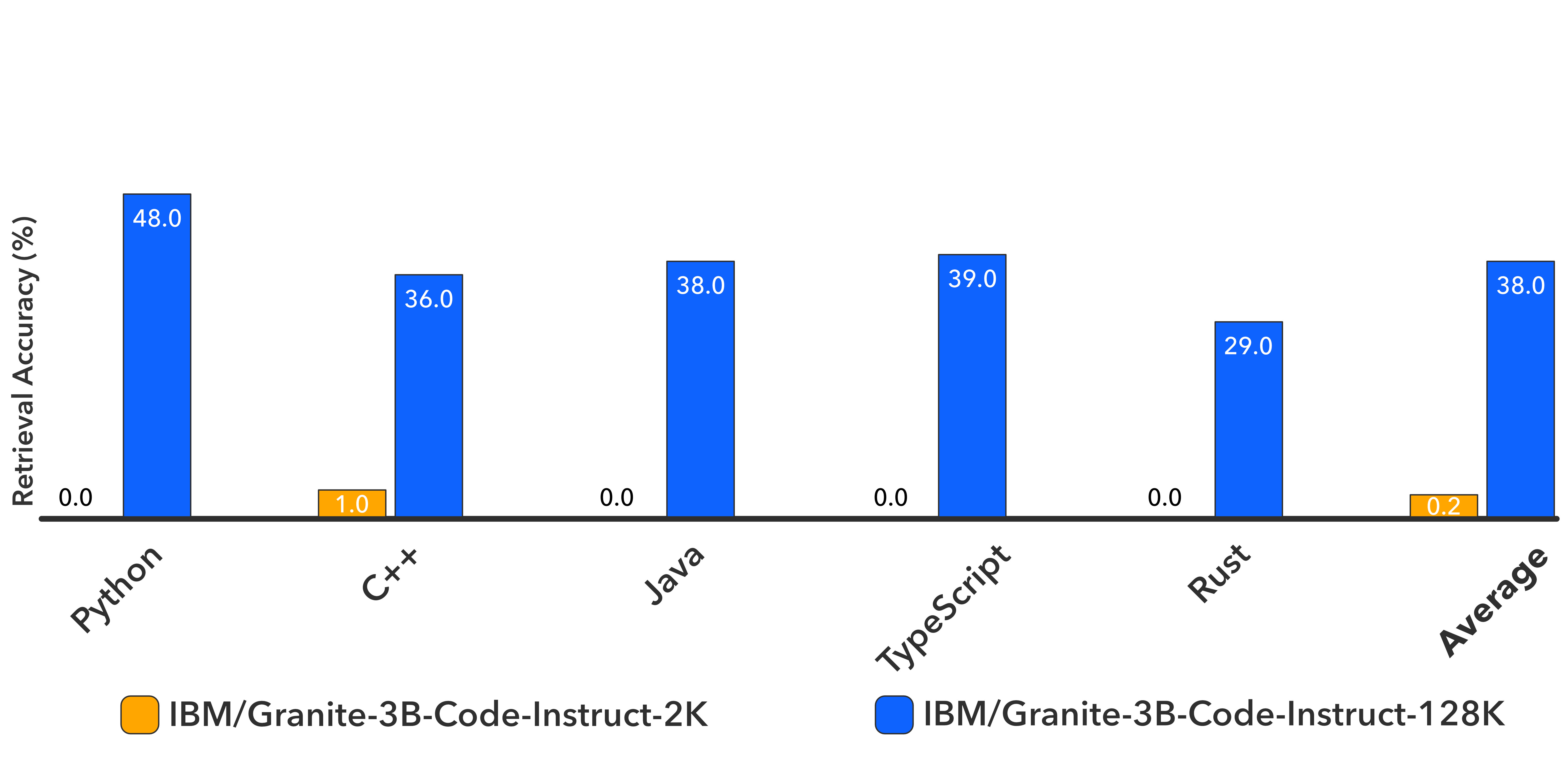}
\includegraphics[width=0.8\textwidth]{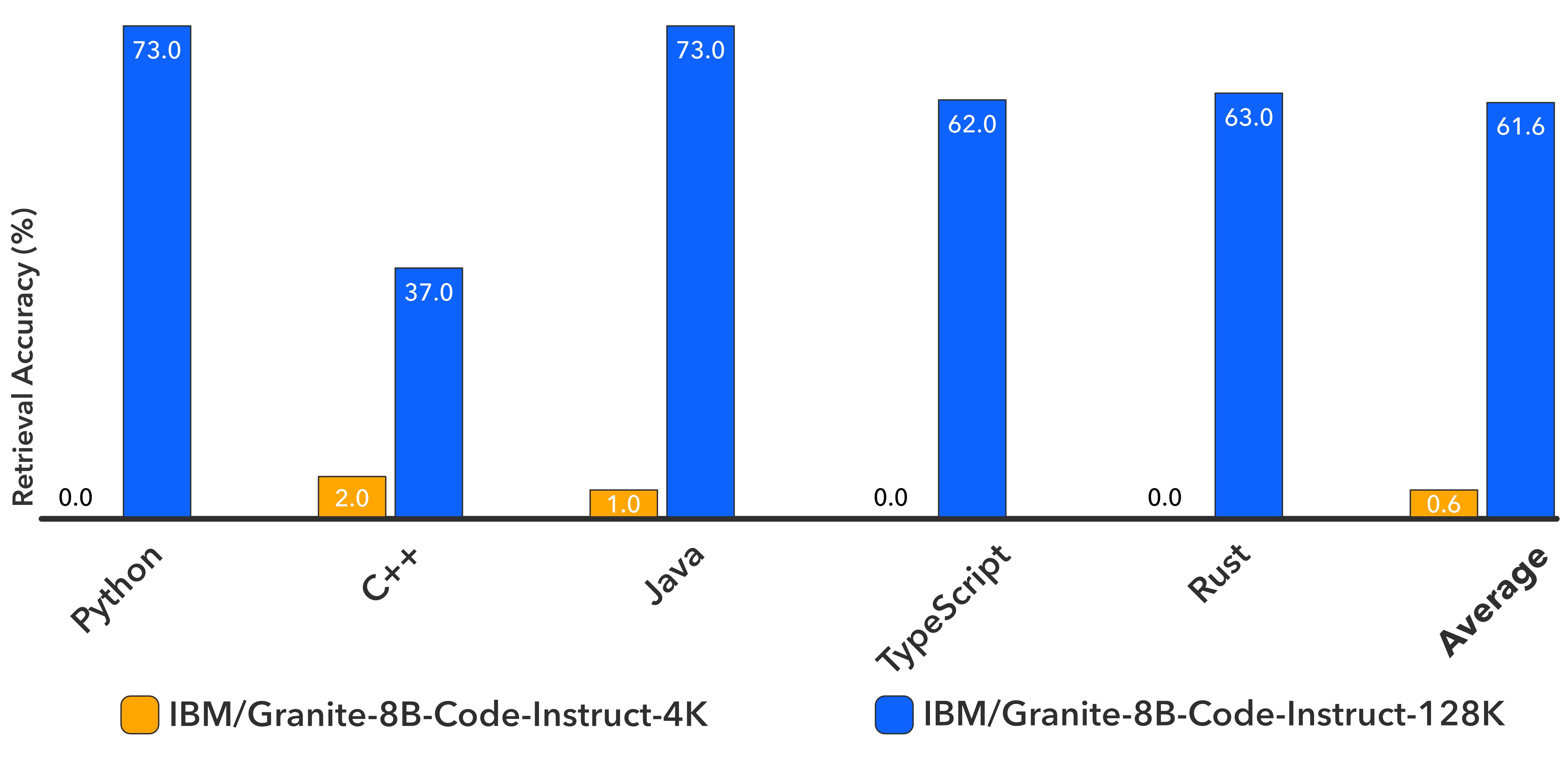}
\caption{Retrieval accuracy of Granite 3B/8B code instruct models before and after scaling to 128K context length on RepoQA benchmark (with a matching threshold of 0.5).}
\label{fig:repoqa-ins}
\end{figure}

Table~\ref{tab:repoqa-ins} compares the performance of long-context instruct models to their short-context counterparts on RepoQA benchmark. As can be seen, our long-context instruct models significantly outperforms short-context versions on all 5 programming languages across different similarity thresholds. As an illustration, figure~\ref{fig:repoqa-ins} demonstrates the difference between short and long-context models at similarity threshold of 0.5, where the performance of both 3B and 8B instruct models with 2K/4K context length support fails to achieve a retrieval accuracy of more than 2\% across 5 languages (on average 0.6\% vs 61.6\% for 8B instruct model). We attribute the improvements to the knowledge learned from newly
introduced synthetic long data for instruction tuning.

\begin{figure}[h]
\begin{center}
\includegraphics[width=\textwidth]{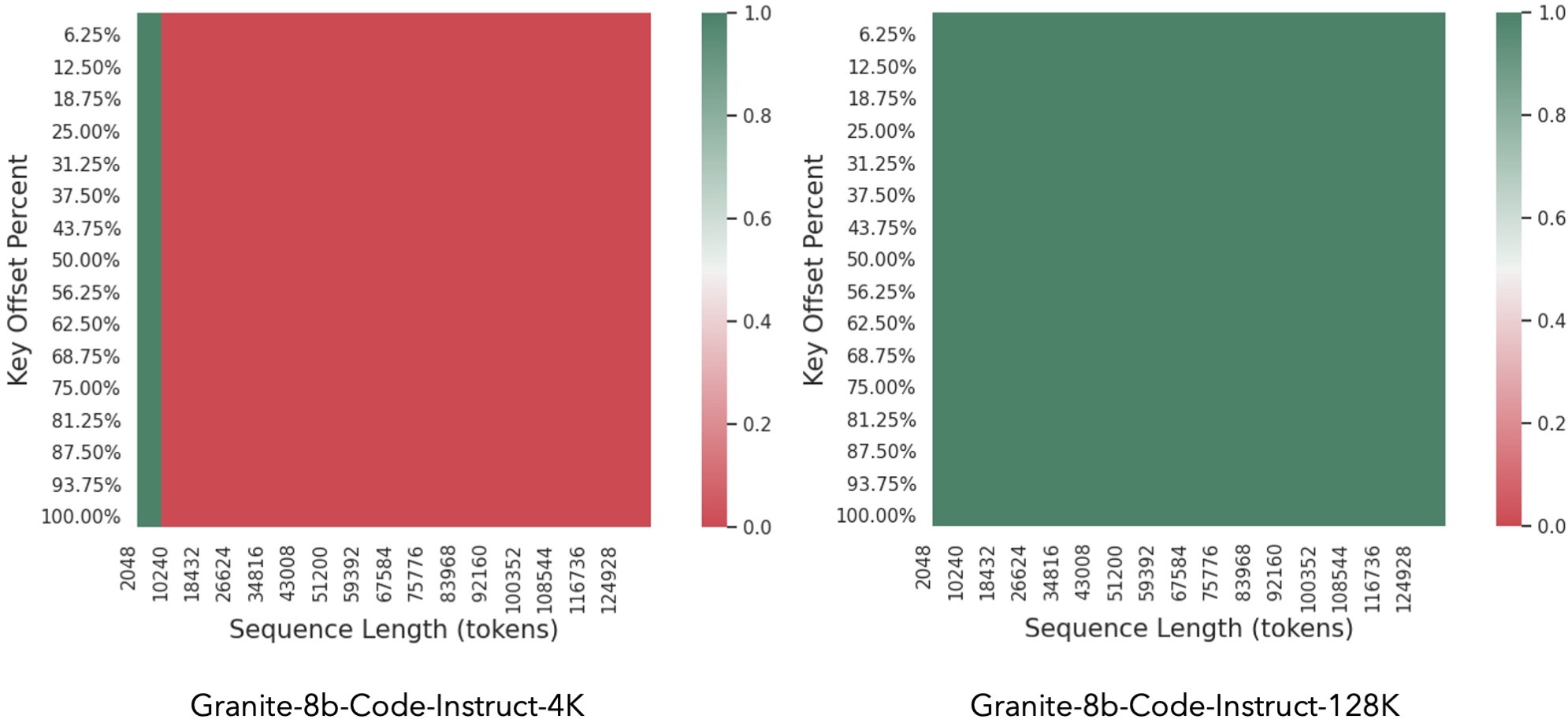}
\end{center} \vspace{-2mm}
\caption{Key retrieval (a.k.a Needle-in-a-Haystack) performance of Granite-8B-Code-Instruct with context scaling. X-axis represents sequence length (tokens) and Y-axis represents key offset percent in retrieval. Best viewed in color.}
\label{fig:key-retr}
\end{figure}

In Figure~\ref{fig:key-retr}, we investigate key retrieval performance of our long-context instruct models on a synthetic benchmark built on top of Python solutions around a key function from  Code Contest finetuning dataset~\citep{alphacoder}. 
Note that this retrieval task is analogous to the famous famous Needle-in-a-Haystack test, albeit tailored to code models.
As can be seen from Figure~\ref{fig:key-retr}, our 8B instruct model before long-context extension only exhibit strong retrieval performance up to 4K length, i.e., on the sequence length they were originally trained on. On the other hand, our context scaling demonstrates a perfect-all-green performance though we tend to view that this level of
retrieval is relatively easy for long-context code LLMs.

\subsection{Short Context Evaluations}

\begin{figure}
\begin{center}
\includegraphics[width=0.6\textwidth]{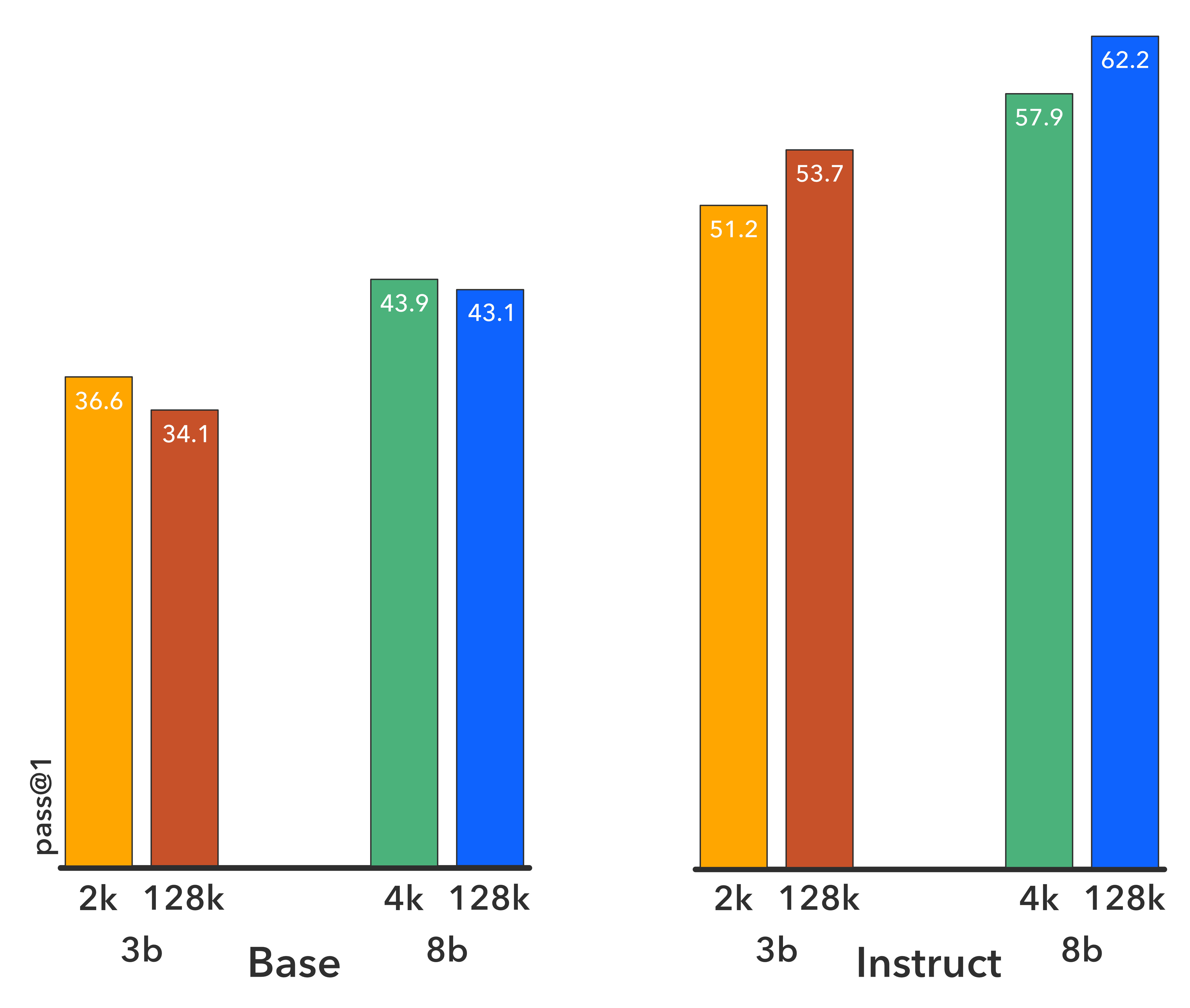}
\end{center} \vspace{-2mm}
\caption{Effect of long-context extension on HumanEval benchmark. While we observe a slight degradation in performance for base models, instruct models see an improvement with long-context scaling, most likely due to our mixing of short-context SFT data with long-context multi-turn synthetic data. Best viewed in color.}
\label{fig:heval-long}
\end{figure}

While our long-context models are very effective on long sequences, we observe that our long-context scaling does not significantly change the short-context generic capability on standard code synthesis benchmarks consisting of short sequences.
Table~\ref{tab:hevalsynthesizelong} summarizes the results on HumanEvalPack, where we find only an average $\sim$1\% degradation for the pass@1 metric on 3B and 8B models respectively. We also test the HumanEval-Python performance in Figure~\ref{fig:heval-long} and observe that long context extension has any noticeable performance degradation. Interestingly, we notice improvements in HumanEval performance of long-context instruct models, which we attribute to our new long-context synthetic data added to instruction tuning. To summarize, while long-context extension comes at a minimal cost for short sequences, we believe this cost is more than offset by the potential of handling long sequences for many real downstream applications. 

\begin{table}[th]
    \caption{Pass@1 performance on HumanEvalPack benchmark~\citep{octopack}. All models are evaluated using greedy decoding with completion format for the base models, and instruction template for the instruction-tuned models.
    }
    \label{tab:hevalsynthesizelong}
    \centering
    \small
    \begin{tabular}{cc|ccc|c}
        \toprule
        \textbf{Model} & \textbf{Prompt} & \textbf{Synthesis} & \textbf{Fix} & \textbf{Explain} & \textbf{Avg.}  \\
        \midrule
        Granite-3b-Code-Base-2K & Completion & 33.0 & 19.5 & 22.2 & 24.9 \\
        Granite-3b-Code-Base-128K & Completion & 30.5 & 19.9 & 22.4 & 24.2 \\
        Granite-8b-Code-Base-4K & Completion & 43.1 & 29.1 & 25.4 & 32.5 \\
        Granite-8b-Code-Base-128K & Completion & 40.2 & 25.2 & 28.2 & 31.2\\
        \midrule
        Granite-3b-Code-Instruct-2K & Instruct & 39.6 & 27.3 & 26.0 & 31.0 \\
        Granite-3b-Code-Instruct-128K & Instruct & 41.4 & 26.2 & 25.1 & 30.9 \\
        Granite-8b-Code-Instruct-4K & Instruct & 49.6 & 40.9 & 40.4 & 43.6 \\
        Granite-8b-Code-Instruct-128K & Instruct & 51.4 & 38.3 & 38.9 & 42.9 \\
        \bottomrule
    \end{tabular}
\end{table}

\section{Conclusion}
\label{sec:conclusion}

We present long-context Granite code models (3B and 8B) that support effective context lengths up to 128K tokens. We perform long context scaling by leveraging a simple yet effective strategy consisting of a lightweight continual pretraining followed by instruction tuning on a mix of short and long-context data. Our long-context models demonstrate much superior performance compared to their short-context counterparts without significantly affecting the short-context generic capability. We believe that given our current results,  methods to enable even longer context length and circumvent the quadratic computational complexity of attention computation will continue to further evolve \citep{gu2023mamba}. We plan to continuously release updates to these models to improve their performance and bringing the best of breed approaches to IBM Granite Family.
\section*{Acknowledgments}

We would like sincerely thank IBM Research leaders - Dario Gil, Sriram Raghavan, Mukesh Khare, Danny Barnett, Talia Gershon, Priya Nagpurkar, Nicholas Fuller for their support. Thanks and acknowledgement to Michele Merler, Shivdeep Singh, Manish Sethi, Pengyuan Li, Kun-Lung Wu, Syed Zawad, Andrew Coleman, Matthew White, Mark Lewis, Raju Pavuluri, Boris Lublinsky, Maximilien de Bayser, Ibrahim Abdelaziz, Kinjal Basu, Mayank Agarwal, Yi Zhou, Chris Johnson, Aanchal Goyal, Yousaf Shah, Petros Zerfos, Heiko Ludwig, Asim Munawar, Maxwell Crouse, Pavan Kapanipathi, Shweta Salaria, Bob Calio, Sophia Wen, Seetharami Seelam, Brian Belgodere, Carlos Fonseca, Colm Malone, Ray Rose, Amith Singhee, Trent Gray-Donald, Xuan Liu, Luis Angel Bathen, Abraham Daniels, Anita Govindjee, Kate Soule, and Lan Hoang.

\newpage
\bibliography{0_main}

\begin{thebibliography}{24}
\providecommand{\natexlab}[1]{#1}
\providecommand{\url}[1]{\texttt{#1}}
\expandafter\ifx\csname urlstyle\endcsname\relax
  \providecommand{\doi}[1]{doi: #1}\else
  \providecommand{\doi}{doi: \begingroup \urlstyle{rm}\Url}\fi

\bibitem[Bai et~al.(2024)Bai, Lv, Zhang, Lyu, Tang, Huang, Du, Liu, Zeng, Hou, Dong, Tang, and Li]{bai2024longbenchbilingualmultitaskbenchmark}
Yushi Bai, Xin Lv, Jiajie Zhang, Hongchang Lyu, Jiankai Tang, Zhidian Huang, Zhengxiao Du, Xiao Liu, Aohan Zeng, Lei Hou, Yuxiao Dong, Jie Tang, and Juanzi Li.
\newblock Longbench: A bilingual, multitask benchmark for long context understanding, 2024.
\newblock URL \url{https://arxiv.org/abs/2308.14508}.

\bibitem[Basu et~al.(2024)Basu, Abdelaziz, Chaudhury, Dan, Crouse, Munawar, Kumaravel, Muthusamy, Kapanipathi, and Lastras]{basu2024api}
Kinjal Basu, Ibrahim Abdelaziz, Subhajit Chaudhury, Soham Dan, Maxwell Crouse, Asim Munawar, Sadhana Kumaravel, Vinod Muthusamy, Pavan Kapanipathi, and Luis~A Lastras.
\newblock Api-blend: A comprehensive corpora for training and benchmarking api llms.
\newblock \emph{arXiv preprint arXiv:2402.15491}, 2024.

\bibitem[{CodeGemma Team} et~al.(2024){CodeGemma Team}, Hartman, Hu, Choquette-Choo, Zhao, Fine, Hui, Shen, Kelley, Howland, Bansal, Vilnis, Wirth, Nguyen, Michel, Choy, Joshi, Kumar, Hashmi, Agrawal, Zuo, Warkentin, and Gong]{codegemma}
{CodeGemma Team}, Ale~Jakse Hartman, Andrea Hu, Christopher~A. Choquette-Choo, Heri Zhao, Jane Fine, Jeffrey Hui, Jingyue Shen, Joe Kelley, Joshua Howland, Kshitij Bansal, Luke Vilnis, Mateo Wirth, Nam Nguyen, Paul Michel, Peter Choy, Pratik Joshi, Ravin Kumar, Sarmad Hashmi, Shubham Agrawal, Siqi Zuo, Tris Warkentin, and Zhitao et~al. Gong.
\newblock Codegemma: Open code models based on gemma.
\newblock 2024.
\newblock URL \url{https://goo.gle/codegemma}.

\bibitem[Fu et~al.(2024)Fu, Panda, Niu, Yue, Hajishirzi, Kim, and Peng]{fu2024data}
Yao Fu, Rameswar Panda, Xinyao Niu, Xiang Yue, Hannaneh Hajishirzi, Yoon Kim, and Hao Peng.
\newblock Data engineering for scaling language models to 128k context.
\newblock \emph{arXiv preprint arXiv:2402.10171}, 2024.

\bibitem[Gu \& Dao(2023)Gu and Dao]{gu2023mamba}
Albert Gu and Tri Dao.
\newblock Mamba: Linear-time sequence modeling with selective state spaces.
\newblock \emph{arXiv preprint arXiv:2312.00752}, 2023.

\bibitem[Guo et~al.(2023)Guo, Xu, Duan, Yin, and McAuley]{guo2023longcoderlongrangepretrainedlanguage}
Daya Guo, Canwen Xu, Nan Duan, Jian Yin, and Julian McAuley.
\newblock Longcoder: A long-range pre-trained language model for code completion, 2023.
\newblock URL \url{https://arxiv.org/abs/2306.14893}.

\bibitem[Lee et~al.(2023)Lee, Hunter, and Ruiz]{platypus2023}
Ariel~N. Lee, Cole~J. Hunter, and Nataniel Ruiz.
\newblock Platypus: Quick, cheap, and powerful refinement of llms.
\newblock 2023.

\bibitem[Li et~al.(2021)Li, Xue, Baranwal, Li, and You]{li2021sequence}
Shenggui Li, Fuzhao Xue, Chaitanya Baranwal, Yongbin Li, and Yang You.
\newblock Sequence parallelism: Long sequence training from system perspective.
\newblock \emph{arXiv preprint arXiv:2105.13120}, 2021.

\bibitem[Li et~al.(2022)Li, Choi, Chung, Kushman, Schrittwieser, Leblond, Eccles, Keeling, Gimeno, Dal~Lago, Hubert, Choy, de~Masson~d’Autume, Babuschkin, Chen, Huang, Welbl, Gowal, Cherepanov, Molloy, Mankowitz, Sutherland~Robson, Kohli, de~Freitas, Kavukcuoglu, and Vinyals]{alphacoder}
Yujia Li, David Choi, Junyoung Chung, Nate Kushman, Julian Schrittwieser, Rémi Leblond, Tom Eccles, James Keeling, Felix Gimeno, Agustin Dal~Lago, Thomas Hubert, Peter Choy, Cyprien de~Masson~d’Autume, Igor Babuschkin, Xinyun Chen, Po-Sen Huang, Johannes Welbl, Sven Gowal, Alexey Cherepanov, James Molloy, Daniel~J. Mankowitz, Esme Sutherland~Robson, Pushmeet Kohli, Nando de~Freitas, Koray Kavukcuoglu, and Oriol Vinyals.
\newblock Competition-level code generation with alphacode.
\newblock \emph{Science}, 378\penalty0 (6624):\penalty0 1092–1097, December 2022.
\newblock ISSN 1095-9203.
\newblock \doi{10.1126/science.abq1158}.
\newblock URL \url{http://dx.doi.org/10.1126/science.abq1158}.

\bibitem[Liu et~al.(2023{\natexlab{a}})Liu, Zaharia, and Abbeel]{liu2023ringattentionblockwisetransformers}
Hao Liu, Matei Zaharia, and Pieter Abbeel.
\newblock Ring attention with blockwise transformers for near-infinite context, 2023{\natexlab{a}}.
\newblock URL \url{https://arxiv.org/abs/2310.01889}.

\bibitem[Liu et~al.(2024)Liu, Tian, Daita, Wei, Ding, Wang, Yang, and Zhang]{repoqa}
Jiawei Liu, Jia~Le Tian, Vijay Daita, Yuxiang Wei, Yifeng Ding, Yuhan~Katherine Wang, Jun Yang, and Lingming Zhang.
\newblock Repoqa: Evaluating long context code understanding.
\newblock \emph{arXiv preprint arXiv:2406.06025}, 2024.

\bibitem[Liu et~al.(2023{\natexlab{b}})Liu, Xu, and McAuley]{liu2023repobench}
Tianyang Liu, Canwen Xu, and Julian McAuley.
\newblock Repobench: Benchmarking repository-level code auto-completion systems.
\newblock \emph{arXiv preprint arXiv:2306.03091}, 2023{\natexlab{b}}.

\bibitem[Liu et~al.(2023{\natexlab{c}})Liu, Xu, and McAuley]{liu2023repobenchbenchmarkingrepositorylevelcode}
Tianyang Liu, Canwen Xu, and Julian McAuley.
\newblock Repobench: Benchmarking repository-level code auto-completion systems, 2023{\natexlab{c}}.
\newblock URL \url{https://arxiv.org/abs/2306.03091}.

\bibitem[Mishra et~al.(2024)Mishra, Stallone, Zhang, Shen, Prasad, Soria, Merler, Selvam, Surendran, Singh, et~al.]{mishra2024granite}
Mayank Mishra, Matt Stallone, Gaoyuan Zhang, Yikang Shen, Aditya Prasad, Adriana~Meza Soria, Michele Merler, Parameswaran Selvam, Saptha Surendran, Shivdeep Singh, et~al.
\newblock Granite code models: A family of open foundation models for code intelligence.
\newblock \emph{arXiv preprint arXiv:2405.04324}, 2024.

\bibitem[Muennighoff et~al.(2023)Muennighoff, Liu, Zebaze, Zheng, Hui, Zhuo, Singh, Tang, von Werra, and Longpre]{octopack}
Niklas Muennighoff, Qian Liu, Armel Zebaze, Qinkai Zheng, Binyuan Hui, Terry~Yue Zhuo, Swayam Singh, Xiangru Tang, Leandro von Werra, and Shayne Longpre.
\newblock Octopack: Instruction tuning code large language models, 2023.

\bibitem[{OpenDevin Team}(2024)]{opendevin2024}
{OpenDevin Team}.
\newblock {OpenDevin: An Open Platform for AI Software Developers as Generalist Agents}.
\newblock \url{https://github.com/OpenDevin/OpenDevin}, 2024.
\newblock Accessed: ENTER THE DATE YOU ACCESSED THE PROJECT.

\bibitem[Pinnaparaju et~al.(2024)Pinnaparaju, Adithyan, Phung, Tow, Baicoianu, Datta, Zhuravinskyi, Mahan, Bellagente, Riquelme, et~al.]{pinnaparaju2024stable}
Nikhil Pinnaparaju, Reshinth Adithyan, Duy Phung, Jonathan Tow, James Baicoianu, Ashish Datta, Maksym Zhuravinskyi, Dakota Mahan, Marco Bellagente, Carlos Riquelme, et~al.
\newblock Stable code technical report.
\newblock \emph{arXiv preprint arXiv:2404.01226}, 2024.

\bibitem[Rozière et~al.(2023)Rozière, Gehring, Gloeckle, Sootla, Gat, Tan, Adi, Liu, Remez, Rapin, Kozhevnikov, Evtimov, Bitton, Bhatt, Ferrer, Grattafiori, Xiong, Défossez, Copet, Azhar, Touvron, Martin, Usunier, Scialom, and Synnaeve]{codellama}
Baptiste Rozière, Jonas Gehring, Fabian Gloeckle, Sten Sootla, Itai Gat, Xiaoqing~Ellen Tan, Yossi Adi, Jingyu Liu, Tal Remez, Jérémy Rapin, Artyom Kozhevnikov, Ivan Evtimov, Joanna Bitton, Manish Bhatt, Cristian~Canton Ferrer, Aaron Grattafiori, Wenhan Xiong, Alexandre Défossez, Jade Copet, Faisal Azhar, Hugo Touvron, Louis Martin, Nicolas Usunier, Thomas Scialom, and Gabriel Synnaeve.
\newblock Code llama: Open foundation models for code, 2023.

\bibitem[Wang et~al.(2023{\natexlab{a}})Wang, Ivison, Dasigi, Hessel, Khot, Chandu, Wadden, MacMillan, Smith, Beltagy, and Hajishirzi]{wang2023farcamelsgoexploring}
Yizhong Wang, Hamish Ivison, Pradeep Dasigi, Jack Hessel, Tushar Khot, Khyathi~Raghavi Chandu, David Wadden, Kelsey MacMillan, Noah~A. Smith, Iz~Beltagy, and Hannaneh Hajishirzi.
\newblock How far can camels go? exploring the state of instruction tuning on open resources, 2023{\natexlab{a}}.
\newblock URL \url{https://arxiv.org/abs/2306.04751}.

\bibitem[Wang et~al.(2023{\natexlab{b}})Wang, Dong, Zeng, Adams, Sreedhar, Egert, Delalleau, Scowcroft, Kant, Swope, and Kuchaiev]{wang2023helpsteer}
Zhilin Wang, Yi~Dong, Jiaqi Zeng, Virginia Adams, Makesh~Narsimhan Sreedhar, Daniel Egert, Olivier Delalleau, Jane~Polak Scowcroft, Neel Kant, Aidan Swope, and Oleksii Kuchaiev.
\newblock Helpsteer: Multi-attribute helpfulness dataset for steerlm, 2023{\natexlab{b}}.

\bibitem[Xiong et~al.(2023)Xiong, Liu, Molybog, Zhang, Bhargava, Hou, Martin, Rungta, Sankararaman, Oguz, et~al.]{xiong2023effective}
Wenhan Xiong, Jingyu Liu, Igor Molybog, Hejia Zhang, Prajjwal Bhargava, Rui Hou, Louis Martin, Rashi Rungta, Karthik~Abinav Sankararaman, Barlas Oguz, et~al.
\newblock Effective long-context scaling of foundation models.
\newblock \emph{arXiv preprint arXiv:2309.16039}, 2023.

\bibitem[Yu et~al.(2023)Yu, Jiang, Shi, Yu, Liu, Zhang, Kwok, Li, Weller, and Liu]{yu2023metamath}
Longhui Yu, Weisen Jiang, Han Shi, Jincheng Yu, Zhengying Liu, Yu~Zhang, James~T Kwok, Zhenguo Li, Adrian Weller, and Weiyang Liu.
\newblock Metamath: Bootstrap your own mathematical questions for large language models.
\newblock \emph{arXiv preprint arXiv:2309.12284}, 2023.

\bibitem[Yu(2023)]{yu2023paraphrasing}
Yijiong Yu.
\newblock " paraphrasing the original text" makes high accuracy long-context qa.
\newblock \emph{arXiv preprint arXiv:2312.11193}, 2023.

\bibitem[Yue et~al.(2023)Yue, Qu, Zhang, Fu, Huang, Sun, Su, and Chen]{yue2023mammoth}
Xiang Yue, Xingwei Qu, Ge~Zhang, Yao Fu, Wenhao Huang, Huan Sun, Yu~Su, and Wenhu Chen.
\newblock Mammoth: Building math generalist models through hybrid instruction tuning.
\newblock \emph{arXiv preprint arXiv:2309.05653}, 2023.

\end{thebibliography}
\bibliographystyle{colm2024_conference}

\newpage
\appendix

\end{document}